\pdfoutput=1

\documentclass{article}
\usepackage{ijcai22}

\usepackage{times}
\usepackage{soul}
\usepackage{url}
\usepackage[hidelinks]{hyperref}
\usepackage[utf8]{inputenc}
\usepackage[small]{caption}
\usepackage{graphicx}
\usepackage{amsmath}
\usepackage{amsthm}
\usepackage{booktabs}
\usepackage{algorithm}
\usepackage{algorithmic}
\usepackage{bbm}
\usepackage{multirow}
\usepackage{makecell}
\usepackage{color}
\title{Video Frame Interpolation Based on Deformable Kernel Region}

\author{
	Haoyue Tian$^1$
	\and
	Pan Gao$^1$ \thanks{Corresponding author.} \and
	Xiaojiang Peng$^2$
	
	\affiliations
	$^1$College of Computer Science and Technology, Nanjing University of Aeronautics and Astronautics\\
	$^2$College of Big Data and Internet, Shenzhen Technology University
	
	\emails
	\{tianhy, pan.gao\}@nuaa.edu.cn,
	xiaojiangp@gmail.com
}

\begin{document}
	
	\maketitle
	
	\begin{abstract}
		
		Video frame interpolation task has recently become more and more prevalent in the computer vision field. At present, a number of researches based on deep learning have achieved great success. Most of them are either based on optical flow information, or  interpolation kernel, or a combination of these two methods. However, these methods have ignored that there are grid restrictions on the position of kernel region during synthesizing each target pixel. These limitations result in that they cannot well adapt to the irregularity of object shape and uncertainty of motion, which may lead to irrelevant reference pixels used for interpolation. In order to solve this problem, we revisit the deformable convolution for video interpolation, which can break the fixed grid restrictions on the kernel region, making the distribution of reference points more suitable for the shape of the object, and thus warp a more accurate interpolation frame. Experiments are conducted on four datasets to demonstrate the superior performance of the proposed model in comparison to the state-of-the-art alternatives.
	\end{abstract}
	
	\vspace{-3mm}
	\section{Introduction}
	
	Video frame interpolation aims to synthesize a new frame between two consecutive frames, which is widely used in video processing, such as video encoding, frame rate conversion, or generating slow motion video \cite{slow2020}. However, the complex information of video content, including irregular shapes of objects, various motion patterns or occlusion issues, etc., poses a major challenge to the authenticity of the interpolated frames. With the extensive development of deep learning in computer vision, more and more studies use deep learning methods to interpolate video frames.
	
	Flow-based method is a common approach in this area, which warps the input reference frames according to the optical flow between the interpolation frame and the left and right reference frames, so as to predict the intermediate frame. Some researches such as \cite{ofe2017} and \cite{f2e2017} have developed several methods based on end-to-end flow estimation models for frame interpolation. However, since the interpolated frame does not really exist, the model cannot predict the accurate optical flow information between the interpolated frame and the reference frame. Therefore, low-quality images are generated when the optical flow is not suitable.
	
	Kernel-based method is proposed by \cite{vfi2017}, which regards pixel interpolation as the convolution of corresponding image blocks in two reference frames. Compared with the flow based method, the kernel based method is more flexible. It uses a deep convolution neural network to estimate spatial adaptive convolution kernel, which unifies motion estimation and pixel synthesis into a single process, which can better deal with challenging frame interpolation scenes. However, the kernel size limits the accuracy of the predicted frame. Specifically, if kernel size is too small, the model cannot handle large moving objects, but if the model is set with a large kernel size, it requires a great amount of memory. Although SepConv \cite{vfis2017} reduces memory consumption, it still cannot solve the problem of motion larger than the pre-defined kernel size.
	
	At present, many researches have developed novel algorithms to simultaneously estimate the flow and compensation kernels with respect to the original reference frames, which can tightly couple the motion estimation and kernel estimation networks together to optimize the immediate frame through an overall network model. On the one hand, compared with the flow based method which relies on simple bilinear coefficients, this method can improve the interpolation accuracy by using data-driven kernel coefficients. On the other hand, the optical flow predicted by the flow estimation model first locates the approximate reference kernel region, which can greatly reduce the kernel size, to obtain higher computational efficiency than the sole kernel based method.
	
	Although these methods have improved the video frame interpolation, they all ignore that the object itself has an irregular shape and the motion trajectory is not fixed. Therefore, the regular convolution kernel shape cannot adapt to various objects and motions. In this paper, inspired by Deformable Convolution (DCN) \cite{DCN2017}, we propose to offset the reference pixels adaptively through the latent features extracted by the network, so that the position of the reference point becomes more practical. In addition, according to Deformable Convolution v2 (DCNv2) \cite{DCN22019}, to synthesize a target pixel, not all pixels within the reference field contribute equally to its response, and therefore we utilize both the kernel coefficient and bilinear coefficient to learn the differences in these contributions. 
	Finally, we propose a pixel synthesis module to adaptively combine the optical flow, the interpolation kernel, and the offset field. 
	
	\begin{figure*}[t]
		
		\centerline{\includegraphics[width=0.85\textwidth]{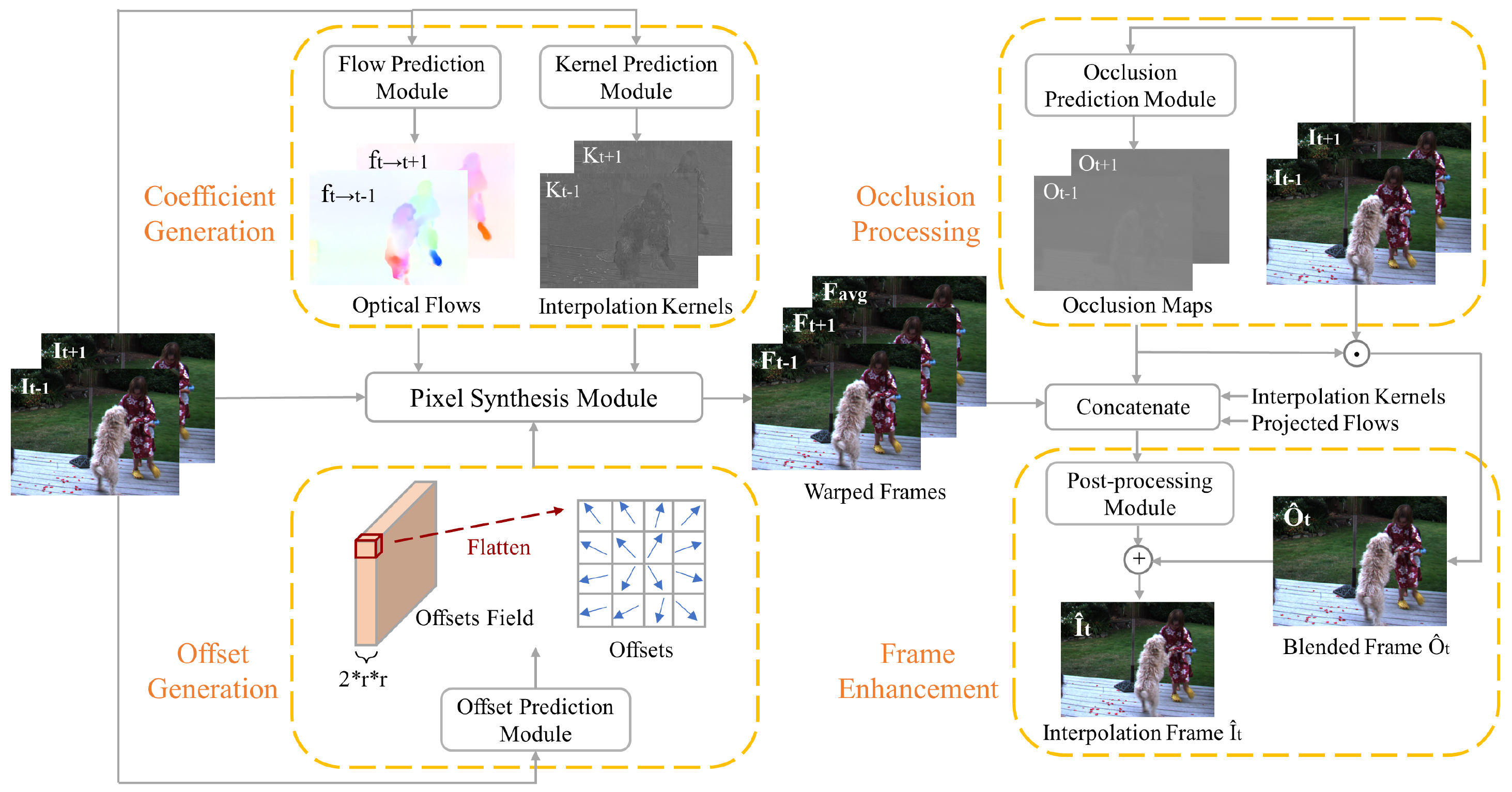}}
		
		\caption{Overall structure of network model. Our model is mainly divided into four submodules including coefficient generation part, offset generation part, occlusion processing part and frame enhancement part. Besides, it also contains a pixel synthesis module used to  adaptively combine optical flow, interpolation kernel, and offset field.}
		\label{Network}
	\end{figure*}  
	
	\section{Related Work}
	
	\subsection{Video Interpolation}
	
	Recently, an increasing number of studies have proved the success of application of deep learning in computer vision, which also inspired various frame interpolation based on deep learning. Super SloMo \cite{SSH2018} utilized two U-Nets to build the network, where one was used to estimate optical flow between two input reference frames, and the other was used to correct the linearly interpolated flow vector, so as to supplement the frame of the video. \cite{qvi2019} proposed a quadratic interpolation algorithm for synthesizing accurate intermediate video frames, which can better simulate the nonlinear motion in the real world by using the acceleration information of the video. Niklaus \emph{et al.} proposed AdaConv \cite{vfi2017} and SepConv \cite{vfis2017} successively, which combined the two steps into a convolution process by convolving the input frame with the spatial adaptive kernel. In addition, SepConv transformed the regular 2D kernel into the 1D kernel, which improves computational efficiency. And it also developed a convolutional neural network that takes in two input frames and estimates pairs of 1D kernels for all pixels simultaneously, which enables the neural network to produce visually pleasing frames.
	
	MEMC-Net \cite{MEMC2019} proposed to exploit motion estimation and motion compensation in a neural network for video frame interpolation. It further proposed an adaptive warping layer for synthesizing new pixels, which can estimate the optical flow and compensation kernel simultaneously. This new warping layer was expected to closely couple the motion estimation and kernel estimation networks. In addition, DAIN \cite{depth2019} utilized the depth information to clearly detect occlusion for video frame interpolation. The bi-directional optical flow and depth map were estimated from the two input frames firstly. Then, instead of simple averaging of flows, the model calculated the contribution of each flow vector according to depth value since multiple flow vectors may be encountered at the same position, which will result in flows with clearer motion boundary.
	
	
	\subsection{Deformable Convolution}
	
	In recent years, Convolution Neural Network (CNN) has made rapid development and progress in the field of computer vision. However, due to the fixed geometric structure in its building module, CNN is inherently limited to model geometric transformation. How to effectively solve the geometric changes in the real world has been a challenge.
	
	In order to handle this problem, \cite{DCN2017} proposed Deformable Convolution Network (DCN) to improve the modeling ability for geometric changes. Specifically, the deformable convolution module first learned offsets based on a parallel network, and then adds these offsets to the position of each sampling point in the convolution kernel, so as to achieve random sampling near the current position without being limited to the regular grid. This made the network more concentrated in the region or target we are interested in.
	
	The problem of DCN was that the introduction of offset module leads to irrelevant context information, which is harmful to the network model. The motivation of Deformable ConvNets v2 (DCNv2) \cite{DCN22019} is to reduce interference information in DCN, which can improve the adaptive capacity of the model to different geometric changes. It added a modulation mechanism in the deformable convolution module, where each sample not only undergoes a learned offset but is also modulated by a learned feature amplitude. Therefore, the network module can change the spatial distribution and relative influence of its samples.
	
	\begin{figure*}[t]
		
		\centerline{\includegraphics[width=0.85\textwidth]{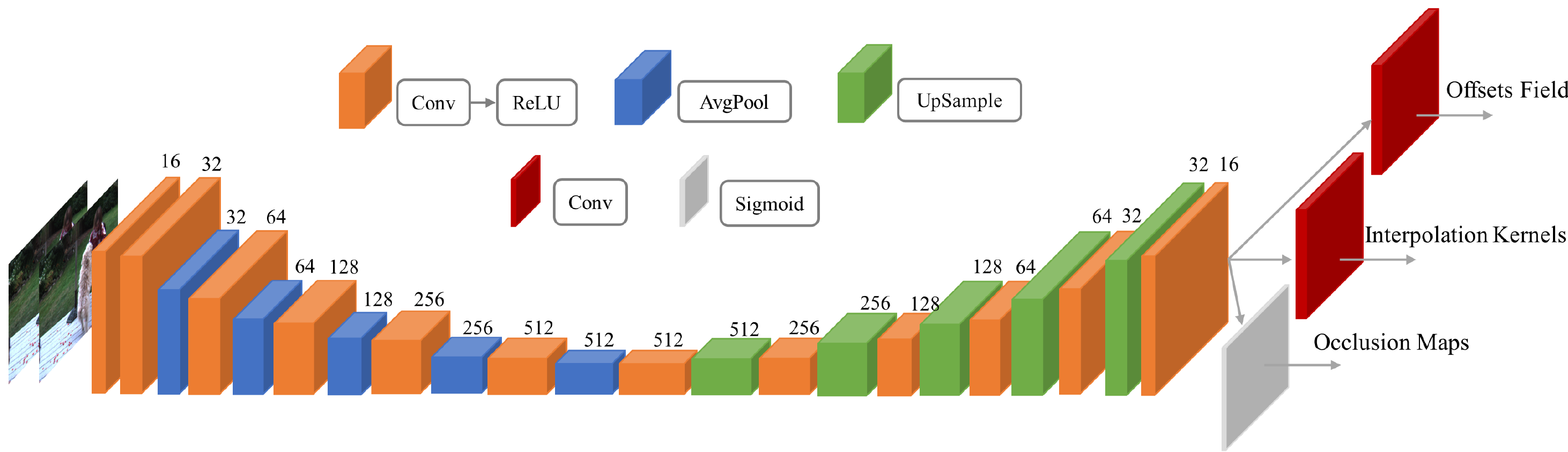}}
		\vspace{-2mm}
		
		\caption{The network structure for kernel prediction, offset prediction  and occlusion prediction modules. For kernel prediction, the network outputs interpolation kernels with  $r*r$ channels; for offset prediction, it outputs offsets field with $2*r*r$ channels; for occlusion prediction, it outputs occlusion maps with 1 channel. In all cases, the network output has the same  spatial resolution with the original frames.}
		\label{subNet}
	\end{figure*}  
	\vspace{-2mm}
	\section{Proposed Approach}
	
	In this section, we introduce the overall structure of our proposed model and the process of pixel synthesis.
	\subsection{Network Architecture}
	
	In this paper, we design a model with deformable kernel region for video frame interpolation. Its overall structure is shown in Figure~\ref{Network}. The input of the model contains two frames, Frame $t-1$, $I_{t-1}$ and Frame $t+1$, $I_{t+1}$, and the purpose is to obtain the intermediate interpolated frame, Frame $t$, between these two frames. Specifically, our model includes four submodules: coefficient generation part, offset generation part, occlusion processing part and frame enhancement part.
	
	\paragraph{Coefficient Generation}
	Inspired by MEMC-Net, this part is designed as a modulation mechanism for pixel synthesis, which consists of bilinear coefficient and kernel coefficient. In particular, we use PWC-Net \cite{pwc2018} to build the optical flow prediction module, which first predicts the optical flow between two reference frames, then obtains the motion vector fields ($f_{t \rightarrow t-1}$ and $f_{t \rightarrow t+1}$) of target frame to reference frames via the flow projection method in \cite{MEMC2019}. Finally, we locate the approximate reference region of each target pixel using $f_{t \rightarrow t-1}$ and $f_{t \rightarrow t+1}$. The kernel prediction module is used to estimate spatially-adaptive convolutional kernels for each target pixel, marked as $K_{t-1}$ and $K_{t+1}$. We consider the U-Net \cite{Unet2015} structure as the basic framework for predicting interpolation kernel coefficients.  The network structure is shown in Figure~\ref{subNet}. We first downsample the input frame several times to extract features through average pooling, and then use bilinear interpolation to upsample the feature maps with the same number of times to reconstruct it. For the kernel prediction module, the interpolation kernel with $r*r$  channels is finally obtained, where $r*r$ represents the number of pixels in the kernel region. This full convolution neural network will also be used to predict offset field and occlusion maps later.
	
	\paragraph{Offset Generation}
	Considering the irregularity of the object and the uncertainty of motion, which were mentioned in DCN and DCNv2 previously, we add an offset generation part to our model. Similarly, we use the network structure shown in Figure~\ref{subNet} as the offset prediction module, and an offset field with $2*r*r$ channels can be obtained, where 2 represents the offset along $x$ direction and $y$ direction, respectively. Then, according to the initial position and offsets of the reference pixels, the kernel region of target pixel is relaxed from the constraints of the regular grid, and finally we finds the more accurate reference position. 
	
	\paragraph{Occlusion Processing}
	
	Due to the relative motion of objects in natural video, there may exist occluded regions between the two reference frames. When this happens, the target pixel become invisible in one of the input frames. Therefore, in order to select effective reference pixels, we use an occlusion prediction module to learn occlusion maps. Since occlusion maps can be understood as the weight of the reference frames, its values are in the range of [0,1]. Therefore, we add sigmoid activation after the basic network structure shown in Figure~\ref{subNet}. The blended frame $\hat{O}_t$ is generated by
	\begin{equation}
		\hat{O}_t = O_{t-1} \cdot I_{t-1} + O_{t+1} \cdot I_{t+1}
	\end{equation}
	where $\cdot$ represents the matrix multiplication of the corresponding elements, $I_{t-1}$ and $I_{t+1}$ are the reference frames.  $O_{t-1}$ is the output of the network and $O_{t+1} = \mathbbm{1} - O_{t-1}$, where $\mathbbm{1}$ is a matrix of ones. Thus, the larger the value of $O_{t-1}$, the better the visibility on $I_{t-1}$. On the contrary, the larger the value of $O_{t+1}$, the better the visibility on $I_{t+1}$. 
	
	\paragraph{Frame Enhancement}
	We also add a post-processing module to enhance the quality of interpolated frame. This module concatenate the warped frames, occlusion maps, interpolation kernels and projected flows as input. Since the input frame and the output frame of our model are quite similar, we make the post-processing module output residual between blended frame $O_t$ and the ground truth. Note that the network structure of the post-processing module is composed of 4 stacked standard convolutions with the filter size of $3 \times 3$, and except for the last layer, each other convolutional layer is followed by an activation function Rectified Linear Unit (ReLU).
	
	
	\begin{figure}[t]
		
		\centerline{\includegraphics[width=0.4\textwidth]{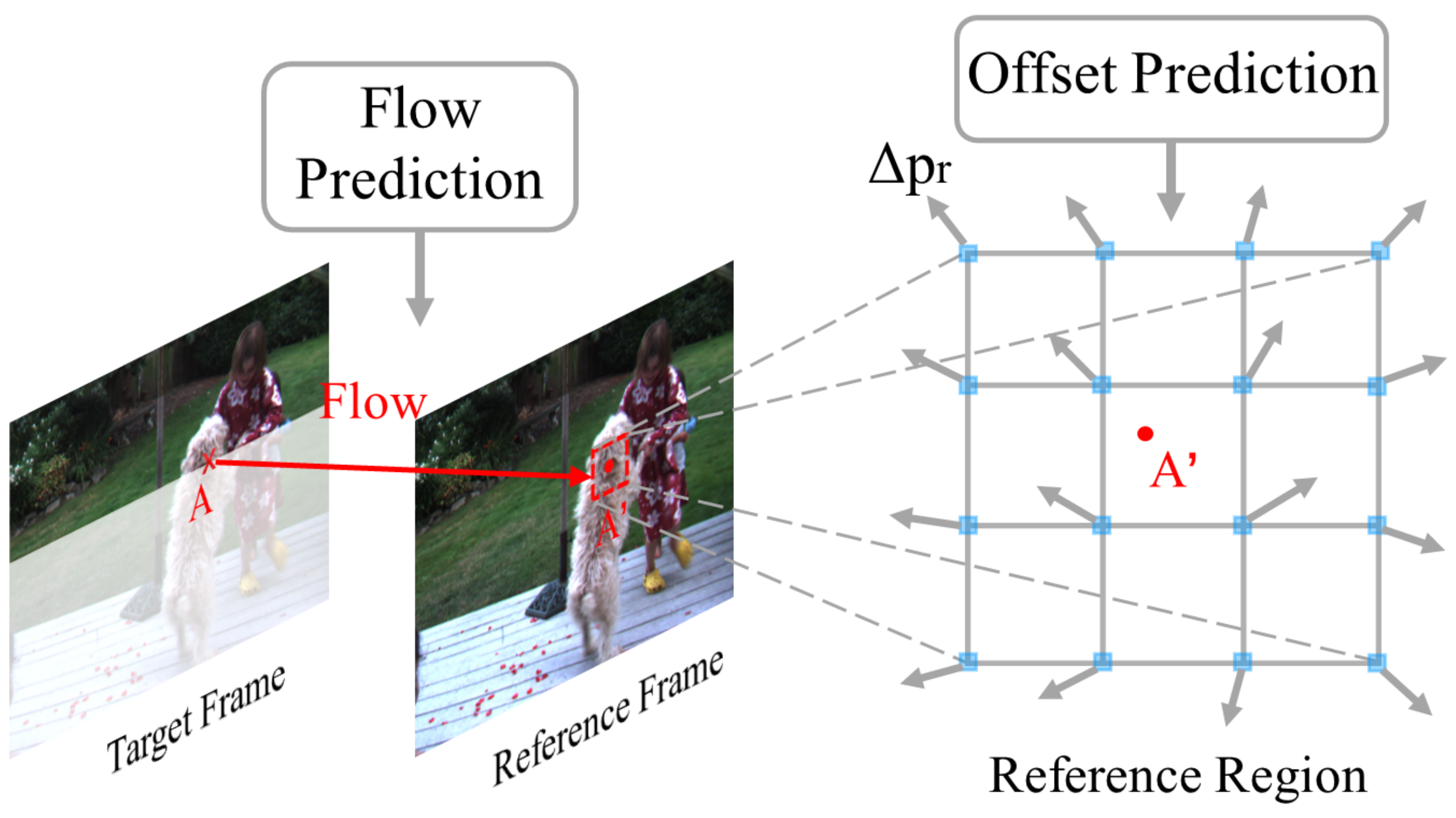}}
		
		\caption{Location of the position of the reference pixels according to the optical flow and offsets.}
		\label{offset}
	\end{figure}  
	
	\begin{figure}[t]
		
		\centerline{\includegraphics[width=0.4\textwidth]{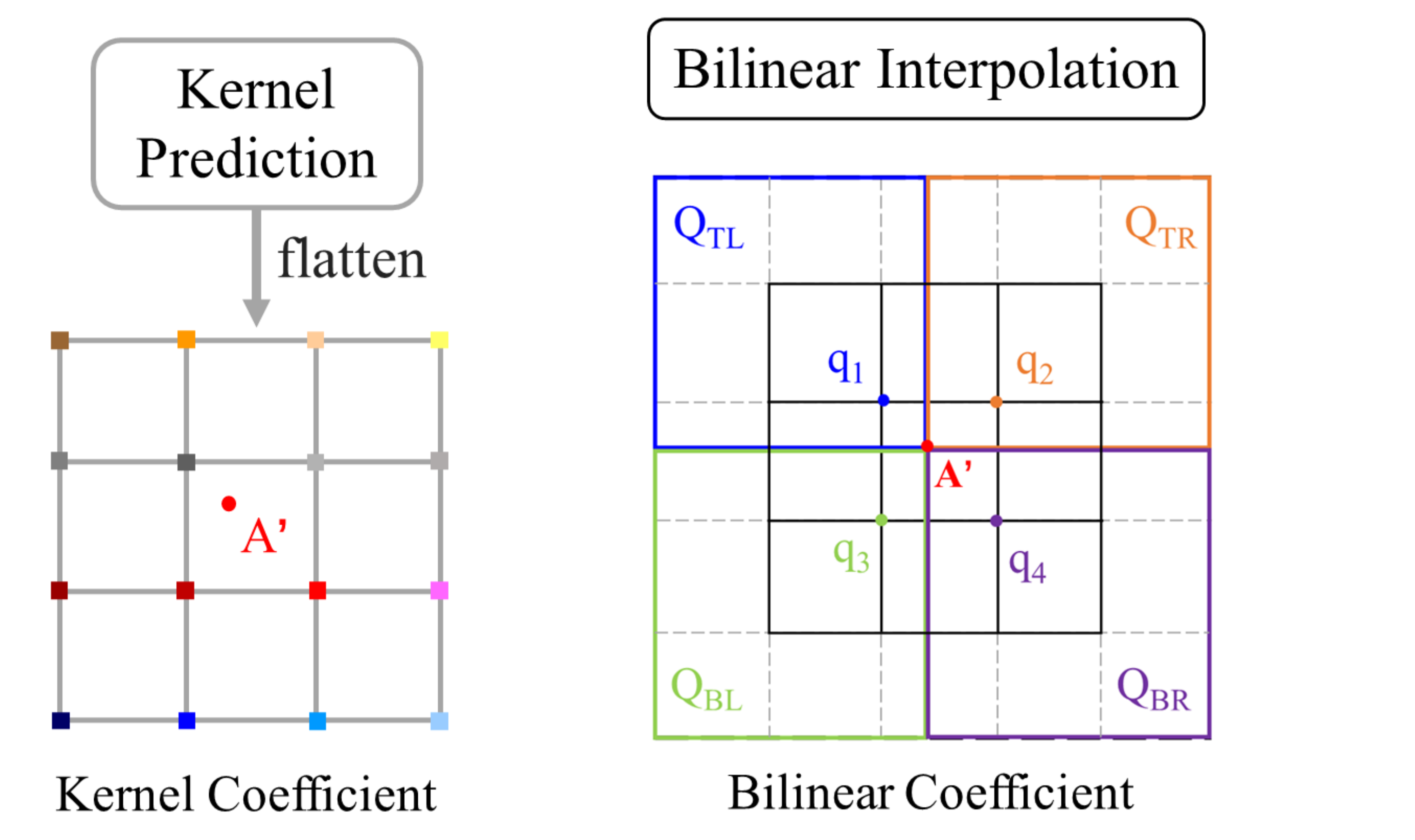}}
		
		\caption{Generation of kernel coefficients and bilinear coefficients.}
		\label{coefficient}
	\end{figure}

	\begin{algorithm}[tb]
		\caption{The Generation of $\hat{I}(A)$}
		\label{Synthesis}
		\textbf{Input}: Reference frames, optical flows, interpolation kernels and offset fields \\
		\textbf{Output}: Warped frames
		
		\begin{algorithmic} 
			\STATE Let $X_L$ and $Y_T$ denote the $x$  and $y$ values of the integer pixel coordinate of the top left corner, respectively.
			\STATE Let $\Delta r$ denote the offset of reference points.
			\STATE Let $A'(x', y')$ denote the point after positioning, as shown in Figure~\ref{coefficient}.
			\FOR {each $i \in [X_L, filterSize + X_L]$ }
			\FOR {each $j \in [Y_T, filterSize + Y_T]$}
			\STATE $OSx = i + \Delta r_x^i$
			\STATE $OSy = j + \Delta r_y^j$
			\IF {$OSx \leq x'$ and $OSy \leq y'$}
			\STATE $ TL $ += $ w_r^{b}(TL) * w_r^{k} * I_B(OSx, OSy) $
			\ENDIF
			\IF {$OSx > x'$ and $OSy \leq y'$}
			\STATE $ TR $ += $ w_r^{b}(TR) * w_r^{k} * I_B(OSx, OSy) $
			\ENDIF
			\IF {$OSx \leq x'$ and $OSy > y'$}
			\STATE $ BL $ += $ w_r^{b}(BL) * w_r^{k} * I_B(OSx, OSy) $
			\ENDIF
			\IF {$OSx > x'$ and $OSy > y'$}
			\STATE $ BR $ += $ w_r^{b}(BR) * w_r^{k} * I_B(OSx, OSy) $
			\ENDIF
			
			\ENDFOR
			\ENDFOR
			\STATE $\hat{I}(A) = TL + TR + BL + BR $
			\RETURN $\hat{I}(A)$
		\end{algorithmic}
	\end{algorithm}
	
	\subsection{Pixel Synthesis}
	The task of the pixel synthesis module is to calculate the value for the target pixel according to the input reference frames, optical flows, interpolation kernels and offsets. First, the module locates a certain position $A'(x', y')$ of the target pixel $A(x, y)$ on the reference frame according to the predicted optical flow, as depicted in Figure~\ref{offset}. Then, we set the size of the kernel region to $4*4$, a total of 16 reference pixels, i.e., R=16, and $p_r$ is regarded as the relative coordinate of each reference pixel
	\begin{equation}
		p_r \in \{(-1, -1), (-1, 0), \dots , (2, 2)\} 
	\end{equation}
	Note that $(0, 0)$ is the nearest integer pixel position on the top-left of $A'$. Next, we add the learned offsets $\Delta p_r$ to the position of initial reference point to get the final adaptive reference point. Since these offsets are all fractional, we obtain the pixel value through bilinear interpolation.
	
	In addition, as shown in Figure~\ref{coefficient}, we flatten the learned interpolation kernel by channel to obtain the kernel coefficients $w_r^k$ of the target pixel. Besides, according to the principle of bilinear interpolation, we calculate the coefficients of $q_1, q_2, q_3, q_4$. Then, due to offset consideration,  we expand the range to the $Q_{TL}, Q_{TR}, Q_{BL}, Q_{BR}$ four parts as the kernel regions, and each part enlarges the original kernel region by one-pixel width along $x$ and $y$ directions. The bilinear coefficient $w_r^b$ of each part is set to the same as the correspoding $q_i$. and the detailed calculation process is as follows:
	\begin{scriptsize}
		\begin{equation}
			\begin{aligned}
				& w_r^{b}(TL) = [1 - \theta(x')][1 - \theta(y')],    & {p_r^x + \Delta p_r^x \leq \theta(x'), p_r^y + \Delta p_r^y \leq \theta(y') }  \\
				& w_r^{b}(TR) = \theta(x')[1 - \theta(y')],   & {p_r^x + \Delta p_r^x > \theta(x'), p_r^y + \Delta p_r^y \leq \theta(y')}   \\
				& w_r^{b}(BL) = [1 - \theta(x')]\theta(y'),   & {p_r^x + \Delta p_r^x \leq \theta(x'), p_r^y + \Delta p_r^y > \theta(y')}   \\
				& w_r^{b}(BR) = \theta(x')\theta(y'),   & {p_r^x + \Delta  p_r^x > \theta(x'), p_r^y + \Delta p_r^y > \theta(y')}
			\end{aligned}
		\end{equation}
	\end{scriptsize}
	where $\theta(x') = x' - \lfloor x' \rfloor$ and $\theta(y') = y' - \lfloor y' \rfloor$, $p_r^x + \Delta p_r^x$ represents the final relative position in the X-axis direction, which is obtained by adding the offset to the integer position of the reference point, and we determine which bilinear coefficient to use according to the final adaptive position. 
	The synthesis process of target pixels $\hat{I}(A)$ can be expressed as:
	\begin{equation}
		\hat{I}(A) = \sum_{r=1}^{R} w_r^b w_r^k \cdot I_B(A + \lfloor f(A) \rfloor + p_r + \Delta p_r),
	\end{equation}
	where $I_B$ represents bilinear interpolation, and $f(A)$ is the optical flow predicted at point A. The calculation process for a target pixel is shown in Algorithm \ref{Synthesis}.

	\begin{table*}
		\centering
		\begin{tabular}{lccccccc}
			\toprule
			\multirow{2}{*}{Methods} & \multirow{2}{*}{\makecell*[c]{Parameters \\ (million)}} & \multicolumn{2}{c}{Vimeo90K}  & \multicolumn{2}{c}{UCF101} & \multicolumn{2}{c}{Davis480p}\\
			\cmidrule{3-8}
			& & PSNR & SSIM & PSNR & SSIM & PSNR & SSIM \\
			\midrule
			CyclicGen & 19.8 &    33.97&     0.9536&   34.99&   0.9578&    25.36& 0.7356\\
			SepConv-$\emph{L}_{f}$& 21.6 &     33.47     &0.9515      &  34.92&  0.9589& 26.40& 0.7549 \\
			SepConv-$\emph{L}_{1}$& 21.6 &     33.80&    0.9555 &        35.12& 0.9612& 26.82& 0.7749\\
			AdaCoF& 21.8 &        \textcolor{blue}{\underline{34.38}}&   0.9562 &        33.93&  0.9421&  26.72&  0.7687 \\
			MEMC& 67.2 &        34.20&   \textcolor{blue}{\underline{0.959}} &        35.16& \textcolor{blue}{\underline{0.9632}}& 27.27& 0.8121\\
			DAIN& 24.0 &        34.04&   0.9581 &        \textcolor{blue}{\underline{35.26}}& 0.963& \textcolor{blue}{\underline{27.31}}& \textcolor{blue}{\underline{0.8148}}\\
			R-SepConv& 13.6 &        34.31&   0.9543 &        35.01& 0.9601& 26.70& 0.7618\\
			Ours& 31.2&        \textcolor{red}{34.52}&   \textcolor{red}{0.9612} &        \textcolor{red}{35.50}& \textcolor{red}{0.9647}& \textcolor{red}{27.46}& \textcolor{red}{0.8164}\\
			\bottomrule
		\end{tabular}
		\caption{Quantitative comparisons on the Vimeo90K, UCF101 and Davis480p. The numbers marked \textcolor{red}{red} and \textcolor{blue}{\underline{blue}} indicate the best and second best image quality evaluation respectively. We also compare the number of model parameters. Our method has the highest PSNR and SSIM on these three datasets.}
		\label{compare}
	\end{table*}
	
	\begin{table*}
		\centering
		\scalebox{0.8}{
			\begin{tabular}{lcccccccccccc}
				\toprule
				& \multicolumn{3}{c}{AdaCoF} & \multicolumn{3}{c}{R-SepConv}  & \multicolumn{3}{c}{DAIN} & \multicolumn{3}{c}{Ours}\\
				\cmidrule{2-13}
				& PSNR & SSIM & Runtime & PSNR & SSIM & Runtime & PSNR & SSIM & Runtime  & PSNR & SSIM & Runtime \\
				\midrule
				Temple3 & 19.75&    \textcolor{red}{0.8938}&     0.369&  \textcolor{red}{19.80} &   \textcolor{blue}{\underline{0.8913}}&    0.218& 19.05 & 0.876& 0.0983& \textcolor{blue}{\underline{19.78}}& 0.8797& 0.0994\\
				
				Cave4& 24.33&    0.6502 &   0.5433 &       24.77& 0.6726& 0.239& \textcolor{blue}{\underline{24.97}}& \textcolor{blue}{\underline{0.7232}}& 0.0974& \textcolor{red}{25.09}& \textcolor{red}{0.7286}& 0.099\\
				Bandage1& 27.32&    \textcolor{red}{0.8677}&  0.4878 &   \textcolor{blue}{\underline{27.36}}&  0.8644&  0.2261&  26.88 & 0.8572& 0.0976 & \textcolor{red}{27.70}& \textcolor{blue}{\underline{0.8665}}& 0.0975 \\
				Temple2& 26.64&    0.9265&  0.4946 &      \textcolor{red}{28.09}& \textcolor{red}{0.9471}& 0.2317& 27.47& 0.9323& 0.0978& \textcolor{blue}{\underline{27.76}}& \textcolor{blue}{\underline{0.9325}}& 0.0974\\
				Cave2& 25.72&      0.5615&   0.5278 &      27.01& 0.6549& 0.2197& \textcolor{blue}{\underline{28.39}} & \textcolor{blue}{\underline{0.7407}}& 0.0978 & \textcolor{red}{28.48} & \textcolor{red}{0.7462} & 0.0976 \\
				Wall& \textcolor{blue}{\underline{32.24}}&   0.8856&   0.4523 &  31.99& 0.8683& 0.22& 32.21& \textcolor{blue}{\underline{0.8878}}& 0.097& \textcolor{red}{32.43}& \textcolor{red}{0.8879}& 0.0969\\
				P\_Shaman1& 33.83&     0.9493&  0.4964 &   34.01& 0.9487& 0.2245& \textcolor{blue}{\underline{35.93}}& \textcolor{blue}{\underline{0.9609}}& 0.0983& \textcolor{red}{36.07}& \textcolor{red}{0.9674}& 0.0984\\
				\cmidrule{1-13}
				Average & 27.12 & 0.8192 & 0.4816 & 27.58 & 0.8353 & 0.2256 & \textcolor{blue}{\underline{27.86}} & \textcolor{blue}{\underline{0.8540}} & 0.0977 & \textcolor{red}{28.17} & \textcolor{red}{0.8584} & 0.0981  \\
				\bottomrule
			\end{tabular}
		}
		\caption{Quantitative comparison of some video sequences in MPI-Sintel dataset. The numbers marked \textcolor{red}{red} and \textcolor{blue}{\underline{blue}} indicate the best and second best image quality evaluation respectively. Our method can achieve higher PSNR and SSIM with relatively less runtime.}
		\label{MSI}
	\end{table*}
	
	It has been proved in \cite{MEMC2019} that bilinear coefficients and interpolation coefficients can be obtained by back-propagating their gradients to the flow estimation network and the kernel estimation network, respectively. Therefore, we only prove that the offset estimation is differentiable. For convenience, we set $q = A + \lfloor f(A) \rfloor + p_r + \Delta p_r$ and $(u(q),v(q))$ represents the coordinate of point $q$. We take the offset on the horizontal component ($\Delta u$) of the current reference point as an example, and the derivative with respect to the offset field is calculated as follows:
	\begin{equation}
		\frac{\partial \hat{I}(A)}{\partial \Delta u} = w_r^b \cdot w_r^k \cdot \frac{\partial I_B(q)}{\partial \Delta u},
	\end{equation}
	where 
	\begin{equation}
		\frac{\partial I_B(q)}{\partial \Delta u} = \varphi_{TL}^u \cdot I_{TL} + \varphi_{TR}^u \cdot I_{TR} + \varphi_{BL}^u \cdot I_{BL} + \varphi_{BR}^u \cdot I_{BR},
	\end{equation}
	where $I_{TL}, I_{TR}, I_{BL}, I_{BR}$ represent the values of four integer pixels around the current fractional pixel, and further
	\begin{equation}
		\varphi_m^u = 
		\begin{cases}
			- [1 - \lambda(v)],    & m = TL,  \\
			[1 - \lambda(v)],   & m = TR,   \\
			- \lambda(v),   & m = BL,   \\
			\lambda(v),   & m = BR,
		\end{cases}
	\end{equation}
	Note that here $\lambda(v) = v(q) - \lfloor v(q) \rfloor$. We only calculate the derivative in the horizontal component, and the calculation method in the vertical component is similar.
	
	\begin{figure*}[t]
		
		\centerline{\includegraphics[width=0.95\textwidth]{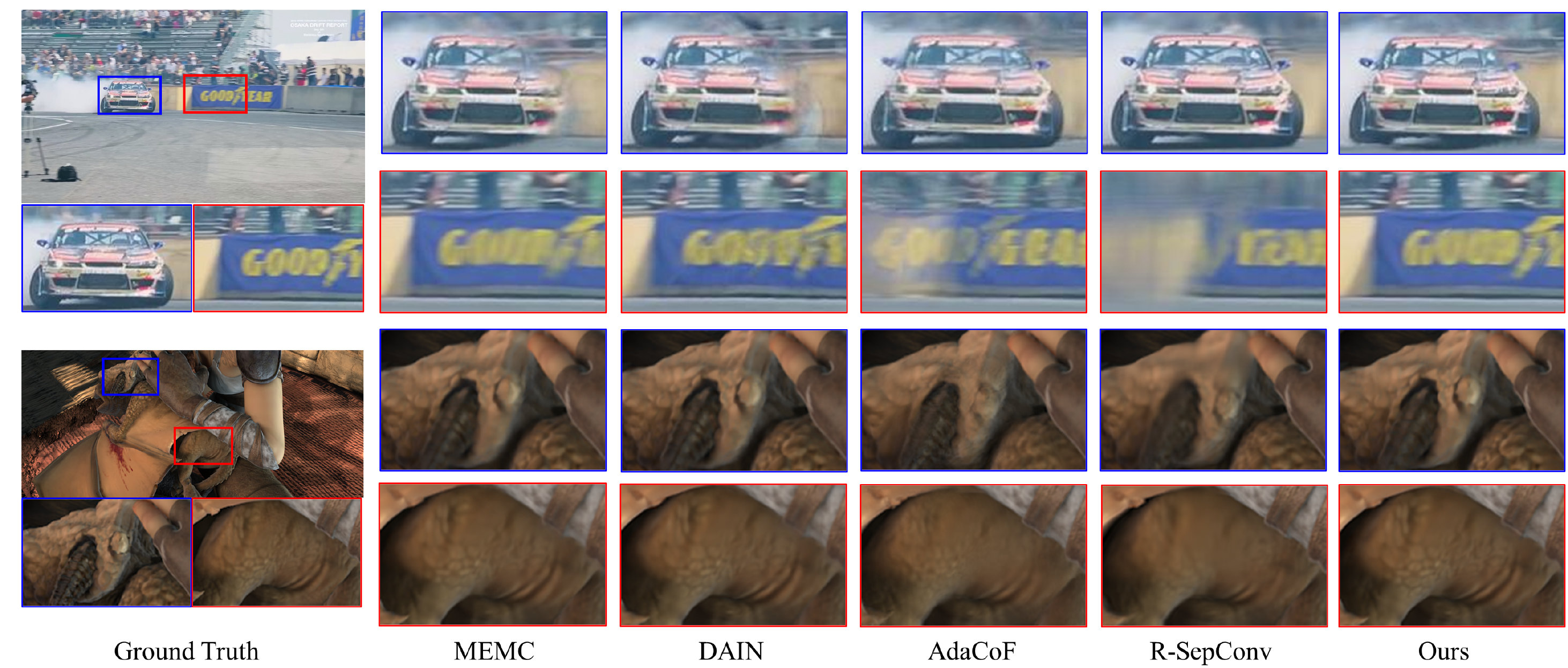}}
		
		\caption{Visual comparison of some sample sequences. The first and second rows are from the Davis480p dataset, and the third and fourth rows are from the MPI-Sintel dataset.}
		\label{result1}
		\vspace{-3mm}
	\end{figure*} 
	
	\section{Experimental details}
	
	The total loss function of our model is designed into two parts: one is used to calculate the loss between the average warped frame ($F_{avg} = (F_{t-1} + F_{t+1})/2$) and the ground truth $I_{gt}$, which is called warped loss $\mathcal L_w$, and the other is enhancement loss $\mathcal L_e$, which is used to calculate the loss between the output image after frame quality enhancement $\hat{I}_{t}$  and the ground truth $I_{gt}$. The total loss function can be formulated as:
	\begin{equation}
		\begin{aligned}
			\mathcal L_{total} &= \lambda_w \mathcal L_w + \lambda_e \mathcal L_e \\
			&= \lambda_w \sum_{x} \Phi \left( F_{avg} - I_{gt} \right) + \lambda_e \sum_{x} \Phi \left( \hat{I}_{t} - I_{gt} \right)
		\end{aligned}
	\end{equation}
	where $\lambda_w$ is set to 1 and $\lambda_e$ is set to 0.5. It is well known that optimization based on $\ell_2$ norm will lead to blurry results in most image synthesis tasks, so we use $\ell_1$ norm for the loss. Inspired by \cite{adacof2020}, \cite{vfig2021}, \cite{depth2019}, we utilize the Charbonnier Penalty Function $\Phi (x) = \sqrt{x^2 + \epsilon ^2} $ to smoothly approximate the $\ell_1$ norm and set $\epsilon = 1e - 6$.
	
	
	We adopt the AdaMax optimizer, where $\beta_1$ and $\beta_2$ are set as the default values $0.9$ and $0.999$, respectively. We set the initial learning rate to 0.002, and during training, if the validation loss does not decrease in 3 consecutive epochs, we reduce the learning rate by a factor of $0.2$. We select Viemo90k \cite{vetf2019} as the dataset and divide it into two parts, which are used for training and validating our proposed model respectively. One part contains 64,600 triples as training set, and the other part has 7,824 triples as the validation set, with a resolution of $448 \times 256$ per frame. We regard the middle frame of the triple as the ground truth, and regard the remaining two frames as the input data. In addition, we also reversed the time order of the input sequence for data enhancement. During training, we set the batch size to 3,  and deploy our experiment to RTX 2080Ti. After training for around 100 epochs, the training loss has converged.

	In order to verify the effectiveness of our model, we also evaluate the trained model on the following four datasets. 
	\paragraph{Vimeo90K Test Set} This dataset \cite{vetf2019} is consisted of 3758 video sequences, each of which has three frames. As in the case of Vimeo90k training dataset, the frame resolution is $448 \times 256$, and we utilize the first and third frames of each sequence to synthesize the second one.
	
	\paragraph{UCF101} The UCF101 \cite{ucf2012} dataset is a large-scale  human behavior dataset. It consists of video sequences containing camera motion and cluttered background. We selected 333 triples from it for model test, of which the resolution is $320 \times 240$.
	
	\paragraph{DAVIS480p} This dataset (Densely Annotated VIdeo Segmentation) \cite{Davis2016} is composed of 50 high-quality, full HD video sequences, covering many common video object segmentation challenges, such as occlusion, motion blur and appearance change. We select 50 groups of three consecutive frames as the test data, and the resolution of each frame is $854 \times 480$.
	
	\paragraph{MPI-Sintel} MPI-Sintel \cite{MPI2012} introduces a new optical flow dataset from an open source 3D animation short film \emph{Sintel}, which has some important features such as long sequence, large motion, specular reflection, motion blur, defocus blur and atmospheric effect. We randomly constructed triples from seven sequences for testing to verify the performance of our model under the above features.

	Table \ref{compare} shows the quantitative comparison between our method and other various state-of-the-art, including CyclicGen \cite{Cyclic2019}, SepConv-$L_f$, SepConv-$L_1$ \cite{SepConv2017}, AdaCoF \cite{adacof2020}, MEMC \cite{MEMC2019}, DAIN \cite{depth2019} and R-SepConv \cite{RSepConv2021}. We measured the PSNR, SSIM and the  model parameters of each method on three test datasets. Although our method does not have the least model parameters, on the whole, it can perform best on three datasets with only a small burden.
	
	In order to make our results more convincing, we selected several effective methods to visualize the \emph{drift-turn} sequence in Davis480p. We took the first frame and the third frame as the input of each model to get the second frame, as shown in the first and second rows of Figure \ref{result1}. Since the motion of the object in these three frames is very large, other methods cannot generate high-quality interpolation frames, for example, even the car wheels disappear or the words on the brand become blurred. But we find our method can synthesize large moving objects well.
	
	We also apply several most advanced interpolation methods on MPI-Sintel datasets and compare them with our method. The quantitative comparison results are shown in Table \ref{MSI}. According to the table, our proposed method can achieve higher PSNR (Peak Signal-to-Noise Ratio) and SSIM (Structural
	Similarity) in a relatively short runtime. We also visualized the interpolation frame,   which is derived from a sequence namely bandage1 in the dataset, as shown in the third and fourth rows in Figure \ref{result1}. According to the results, our method also performs well in synthesizing the detailed texture of the object.
	More qualitative results and ablation experiments are provided in supplementary material. 
	
	\vspace{-2mm}
	\section{Conclusion}
	This paper presents a novel method for video interpolation. Due to the diversity of object shape and the uncertainty of motion, it may not be possible to obtain meaningful feature  with a fixed kernel region. Inspired by deformable convolution, we break the limitation of employing fixed grid reference pixels when synthesizing the target pixel, and make the model learn offsets for all reference points of each target pixel. We then locate the position of the final reference points according to the offsets, and finally synthesize the target pixel. Our model was tested on four datasets, and the experimental results show that our method is superior to most competitive algorithms. 

	\bibliographystyle{named}
	\bibliography{ijcai22}
	
	\clearpage 
	\twocolumn[
	\begin{@twocolumnfalse}
		\section*{ \centering{ \LARGE Supplementary Material for the paper ``Video Frame Interpolation Based on Deformable Kernel Region" }}
			\vspace{25mm}
	\end{@twocolumnfalse}
	]

	\section*{Ablation Study}
	
	To verify the effectiveness of our proposed model, we have performed comprehensive ablation experiments to analyze our network structure. We successively removed the kernel prediction module, offset prediction module, occlusion prediction module and post-processing module from the overall network model presented in the main paper, and combined the remaining modules after each individual module removal. We train the various module combinations using the same method, and then tested them on different datasets. The quantitative results are shown in Table~\ref{ablation}. Note that the flow prediction module was not tested in our ablation experiments as the optical flow information has a direct impact on locating the kernel region.
	
		\setcounter{table}{0}
		\begin{table}[h]
		\centering
		\setlength{\tabcolsep}{1mm}{
			\small
			\begin{tabular}{lcccccc}
				\toprule
				\multirow{2}{*}{Methods}  & \multicolumn{2}{c}{Vimeo90K}  & \multicolumn{2}{c}{UCF101} & \multicolumn{2}{c}{Davis480p}\\
				\cmidrule{2-7}
				& PSNR & SSIM & PSNR & SSIM & PSNR & SSIM \\
				\midrule
				F+OF+OC+P &    33.56&     0.9396&   33.79&   0.9421&    26.48& 0.7748\\
				F+K+OC+P &    33.94     &0.9593      &  34.25& 0.9486& 27.17& 0.8108 \\
				F+K+OF+P &     33.96&    \textcolor{blue}{\underline{0.9596}} &       34.28& 0.9516& 27.05& \textcolor{blue}{\underline{0.8159}}\\
				F+K+OF+OC &     \textcolor{blue}{\underline{33.98}} &   0.9575 &       \textcolor{blue}{\underline{34.84}}& \textcolor{blue}{\underline{0.9616}}&  \textcolor{blue}{\underline{27.19}}&  0.8106 \\
				F+K+OF+OC+P &       \textcolor{red}{34.52}&   \textcolor{red}{0.9612} &     \textcolor{red}{35.50}& \textcolor{red}{0.9647}& \textcolor{red}{27.46}& \textcolor{red}{0.8164}\\
				
				\bottomrule
			\end{tabular}
		}
		\caption{Quantitative comparisons on the Vimeo90K, UCF101 and Davis480p. The numbers marked \textcolor{red}{red} and \textcolor{blue}{\underline{blue}} indicate the best and second best image quality evaluation respectively. Among them, F represents flow prediction module, K is kernel prediction module, OC represents occlusion prediction module, OF is offset prediction module, and P represents post-processing module.}
		\label{ablation}
	\end{table}

	For the F+OF+OC+P combination without K, as shown in the first row of the Table~\ref{ablation}, we found that although the model can reasonably transform the position of the reference pixels by the learned offset, the PSNR and SSIM are still low. This is because the weight of each reference pixel has different influence on the target point. Therefore, when the model lacks of	 the kernel coefficient, it will lead to irrelevant reference, resulting in generating the low-quality interpolation frames. By comparing the second row and fifth row of Table~\ref{ablation}, we observe that when the offset prediction module was added, PSNR is increased by 0.58db, 1.25db and 0.29db respectively on Vimeo90k, UCF101 and Davis480p datasets. When we remove the occlusion module or post-processing module, as illustrated in the third row and fourth row in the Table~\ref{ablation}, the image quality would also decrease. For the improvement brought by the occlusion module, this is due to the relative motion of objects in natural video, and there will be occlusion areas between two reference frames. If this module has not been added, the model may select the occluded invalid reference pixels for interpolation. As for the post-processing module, since the blended image usually contains artifacts caused by inaccurate flow estimation or offset field, this module thus plays the role of eliminating these negative effects, which is also very important for improving overall visual quality.

	\section*{Visualization of Experimental Results}
	
	To better demonstrate the effectiveness of the proposed video frame interpolation model, we provide more qualitative results in this section. We selected several sequences from Vimeo90k and Davis480p datasets for testing, and visualized the interpolated frames to perceive the image visual quality difference between our proposed method and recent state-of-the-art methods, as shown in Figure~\ref{p1} $\sim$~\ref{p8} below. Through observation, we found that the interpolation frame generated by our method has better subjective image quality, especially for the video sequences having complex large motion or occlusion.
	
	\setcounter{figure}{0}
	\begin{figure*}[b]
		
		\centerline{\includegraphics[width=0.85\textwidth]{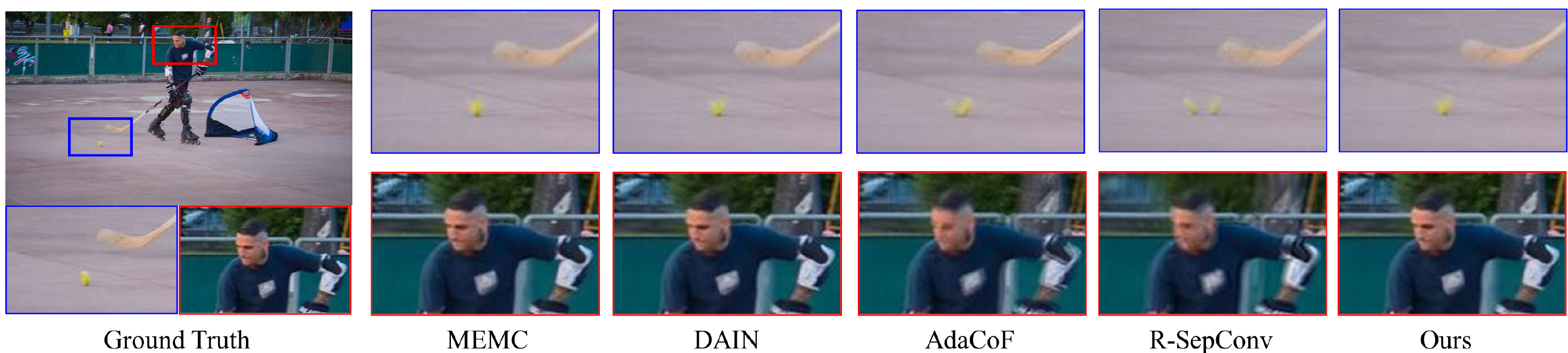}}
		\caption{Visual comparisons on the \textbf{hockey} sequence in Davis480p dataset.}
		\label{p1}
		\vspace{3mm}
	\end{figure*}  
	
	\begin{figure*}[b]
		
		\centerline{\includegraphics[width=0.85\textwidth]{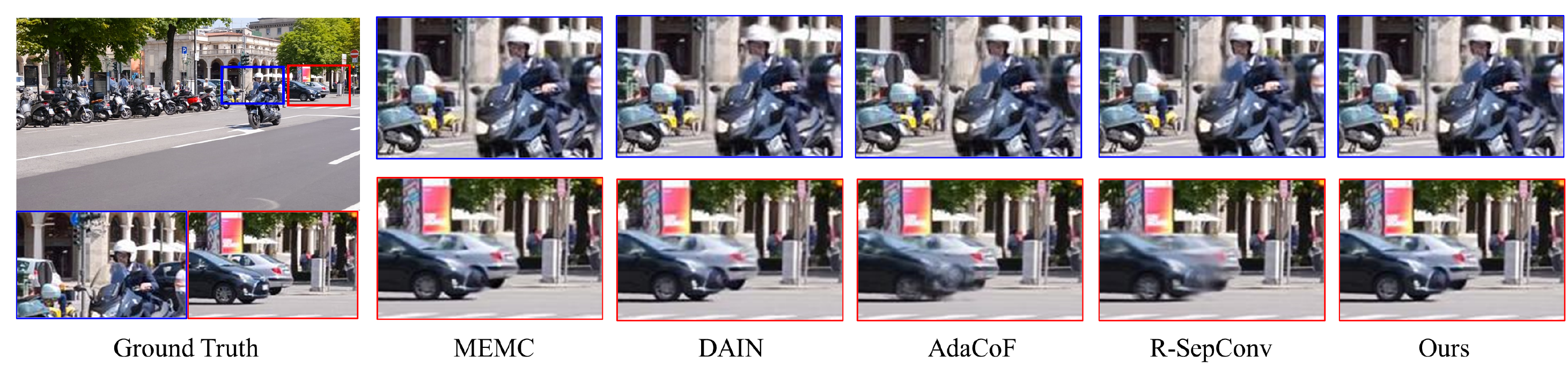}}
		\caption{Visual comparisons on the \textbf{scooter-black} sequence in Davis480p dataset.}
		\label{p2}
		\vspace{3mm}
	\end{figure*}  
	
	\begin{figure*}[t]
		
		\centerline{\includegraphics[width=0.85\textwidth]{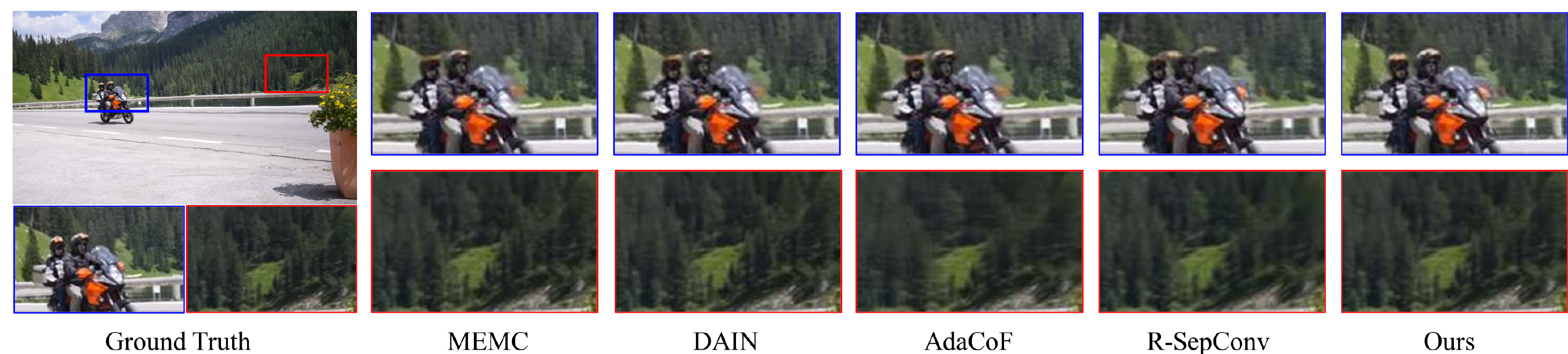}}
		\caption{Visual comparisons on the \textbf{motorbike} sequence in Davis480p dataset.}
		\label{p3}
		\vspace{3mm}
	\end{figure*}
	
	\begin{figure*}[t]
		
		\centerline{\includegraphics[width=0.85\textwidth]{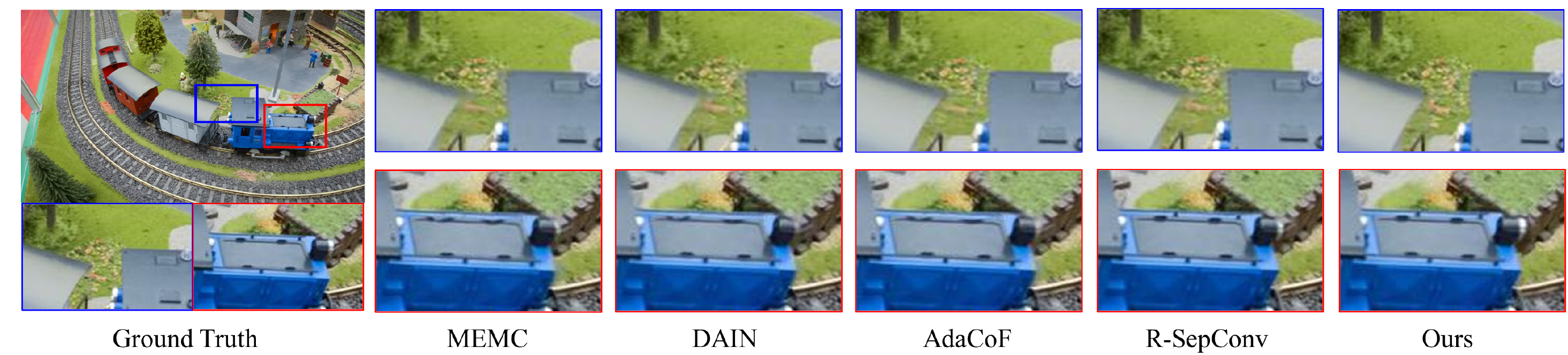}}
		\vspace{3mm}
		\caption{Visual comparisons on the \textbf{train} sequence in Davis480p dataset.}
		\label{p4}
	\end{figure*}
	
	\begin{figure*}[t]
		
		\centerline{\includegraphics[width=0.85\textwidth]{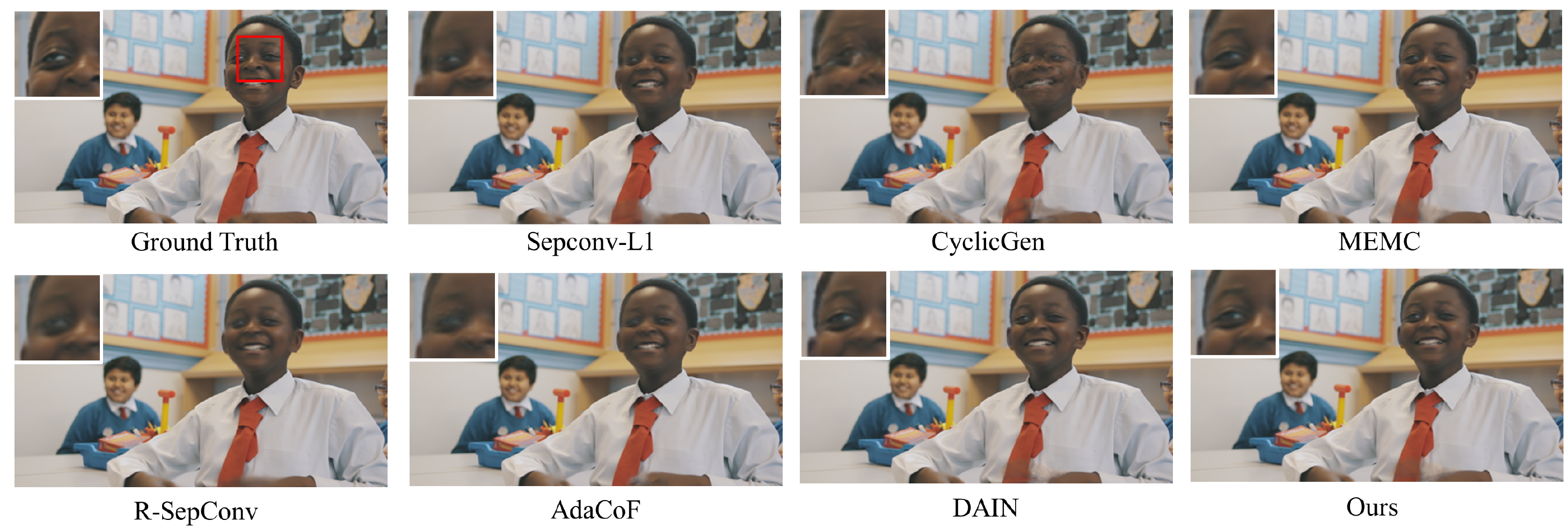}}
		\label{p5}
	\end{figure*}
	
	\begin{figure*}[t]
		
		\centerline{\includegraphics[width=0.85\textwidth]{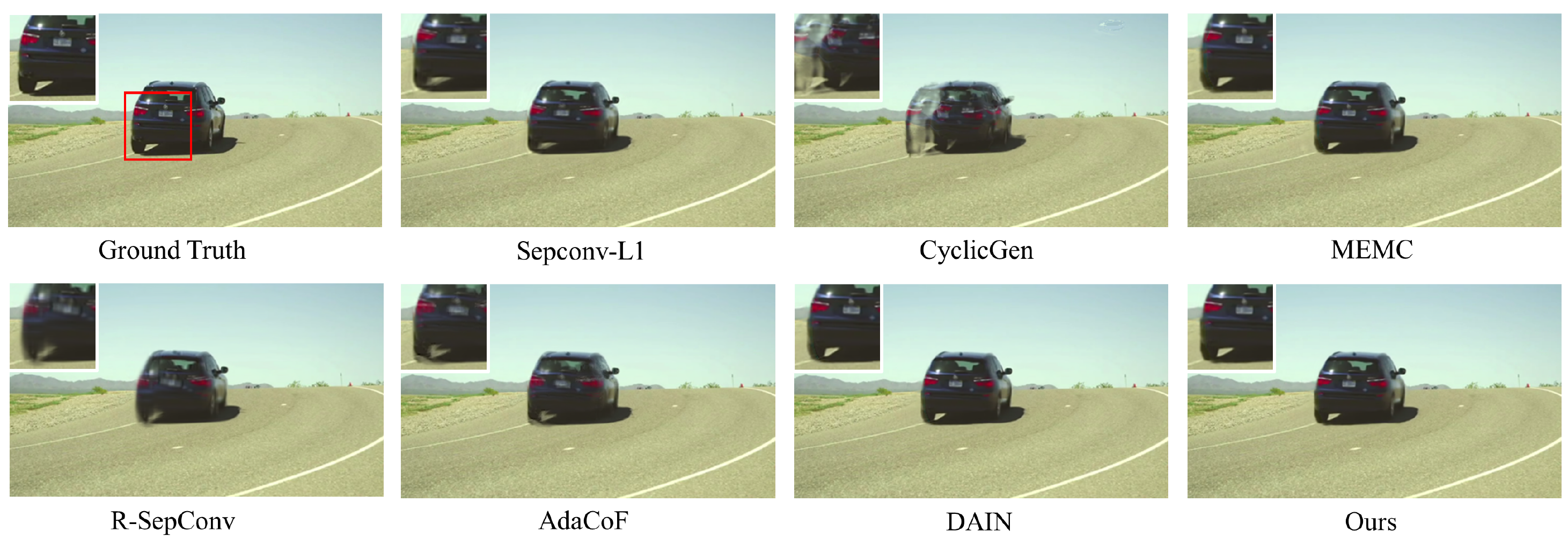}}
		
		\label{p6}
	\end{figure*}
	
	\begin{figure*}[p]
		
		\centerline{\includegraphics[width=0.85\textwidth]{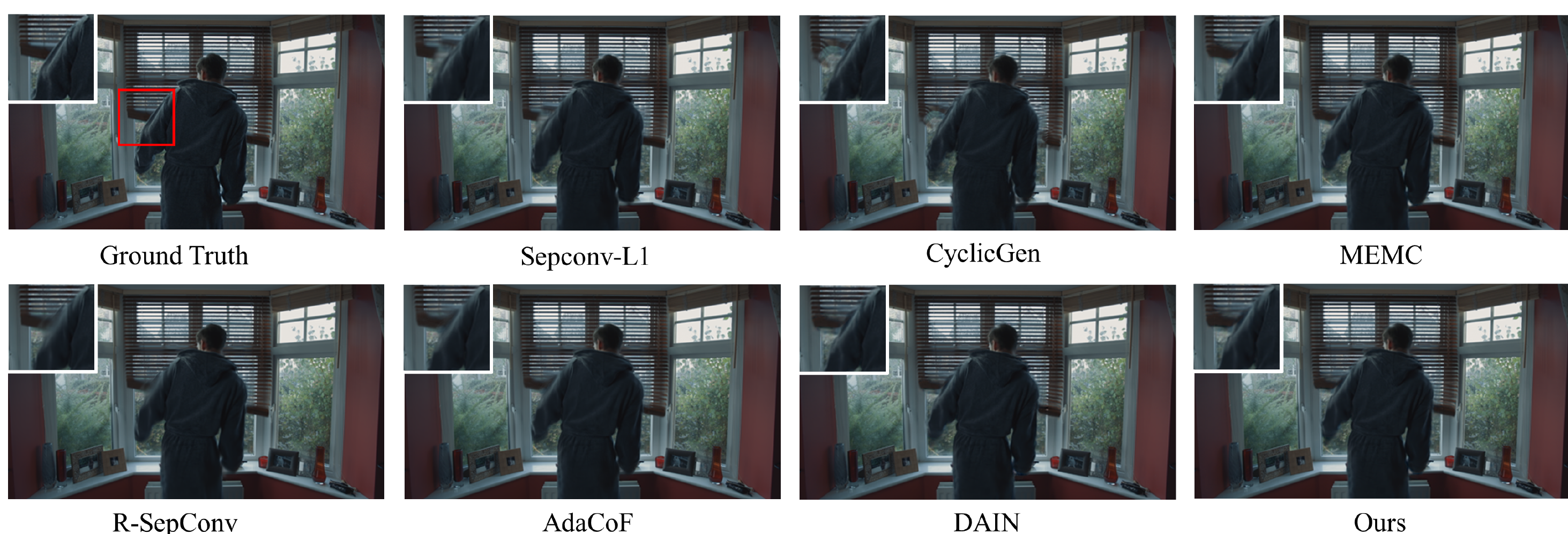}}
		\label{p7}
	\end{figure*}
	
	\begin{figure*}[p]
		
		\centerline{\includegraphics[width=0.85\textwidth]{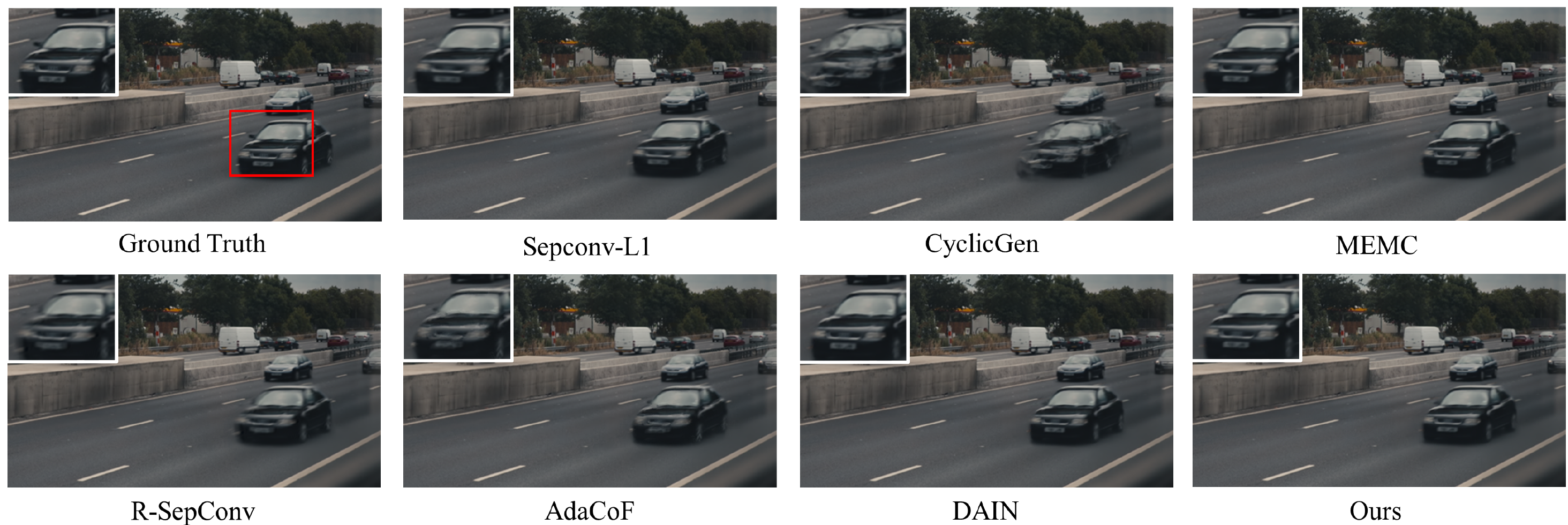}}
		\caption{Visual comparisons of our proposed method with various state-of-the-art methods on the Vimeo90K.}
		\label{p8}
	\end{figure*}

\end{document}